\documentclass[a4paper,11pt]{article}

\usepackage{fourier}
\usepackage{color}
 \usepackage{graphicx}
\usepackage{url}
\usepackage[affil-it]{authblk}
\usepackage{amsmath}
\usepackage{amssymb}
\usepackage{array}
\usepackage{authblk}
\usepackage{wrapfig}
\usepackage{booktabs}
\usepackage{tabularx}
\usepackage[T1]{fontenc}
\usepackage{times}
\usepackage{enumitem}
\usepackage{diagbox}
\usepackage{array}
\usepackage{caption}
\usepackage{microtype}
\usepackage{graphicx}
\usepackage{subcaption}
\usepackage{svg}
\usepackage{indentfirst} % To indent the first paragraph of a section
\usepackage[capitalise]{cleveref} % For better referencing
\usepackage{caption} % For customizing captions

\Crefname{figure}{Fig.}{Figs.}% {<type>}{<singular>}{<plural>}
\Crefname{table}{Tab.}{Tabs.}% {<type>}{<singular>}{<plural>}

\begin{document}

\title{
CrowdSplat: Exploring Gaussian Splatting For Crowd Rendering
}
\author{Xiaohan Sun $^*$}
\author{Yinghan Xu $^*$}
\author{John Dingliana}
\author{Carol O'Sullivan}
\affil{Trinity College Dublin}

\date{}
\maketitle

\def\thefootnote{*}\footnotetext{These authors contributed equally to this work}

\thispagestyle{empty}
\vspace{-20pt}  % Adjust space if needed

\begin{abstract}
We present CrowdSplat, a novel approach 
%leveraging
 that leverages 
3D Gaussian Splatting for real-time, high-quality crowd rendering. Our method utilizes 
%animatable%
3D Gaussian functions to represent animated human characters in diverse poses and outfits, which are extracted from monocular videos. We integrate Level of Detail (LoD) rendering to optimize computational efficiency and quality. The CrowdSplat framework consists of two stages: (1) avatar reconstruction and (2) crowd synthesis. The framework is also optimized in GPU memory usage for scalability. Quantitative and qualitative evaluations show that CrowdSplat achieves good levels of rendering quality, memory efficiency, and computational performance. Through these experiments, we show that CrowdSplat is a viable solution for dynamic, realistic crowd simulation in real-time applications. \textbf{Project Page:} \url{https://github.com/RockyXu66/CrowdSplat}
\end{abstract}


%%%%%%%%%%%%%%%%%%%%%%
\begin{figure}[h!]
    \centering
    \includegraphics[width=\textwidth, trim={0 5.5cm 0 20px},clip]{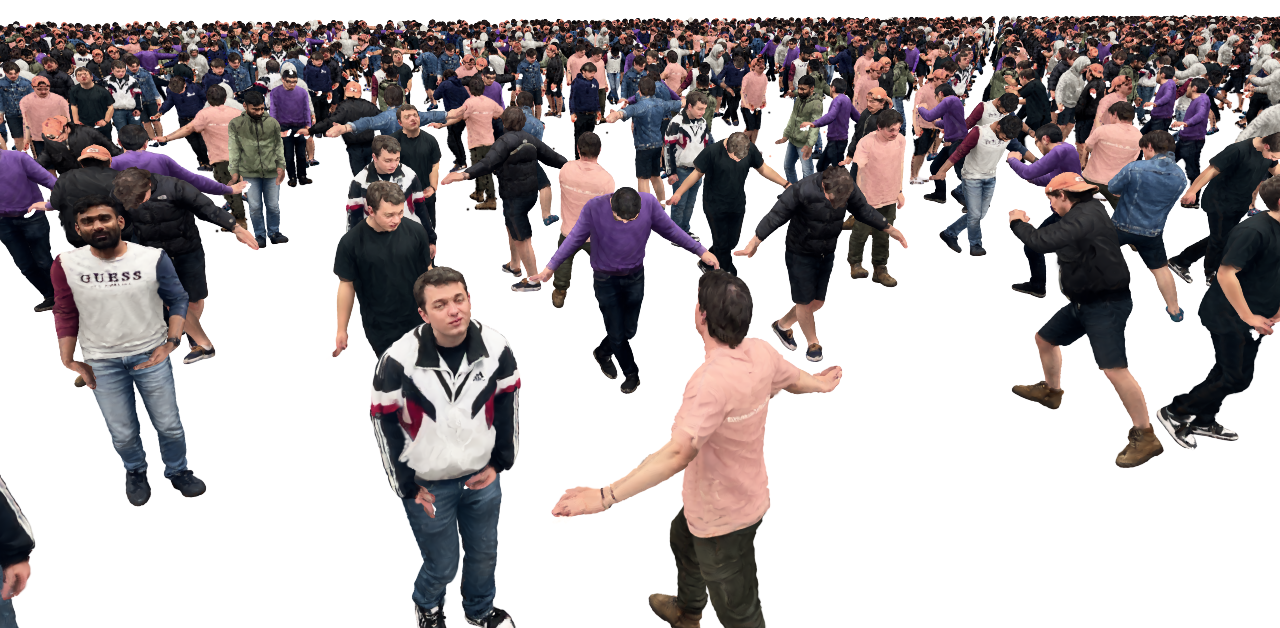}
    \caption{Screenshot of CrowdSplat running at 31 FPS on an RTX4090 with 3,500 animated characters.}
    \label{fig:crowd}
\end{figure}

\vspace{-0.5cm}
\section{Introduction}
In computer graphics, realistic and customizable crowds
%holds significant appeal. Realistic crowd assets are% 
are in demand for applications where visual quality, real-time rendering, and high realism are important (e.g., games and virtual reality). However, generating an animated crowd is challenging as it often demands significant computing resources and manual processing. Traditional crowd rendering and animation methods primarily rely on conventional computer graphics pipelines  and typically require a high degree of data curation and laborious manual editing.

Machine learning has significantly impacted numerous fields, including computer graphics. Innovative approaches like Neural Radiance Fields (NeRF) \cite{N1} and 3D Gaussian Splatting (3DGS) \cite{Kerbl20233DGS} have produced impressive results, marking a new phase in rendering methods. Building upon these advancements, we introduce CrowdSplat, a 3D Gaussian Splatting approach to crowd synthesis.
%rendering. 

Our controllable avatars are reconstructed from monocular videos without manual editing. In conjunction, we use Level Of Detail (LoD) 
%tricks 
methods to enhance the computational efficiency of crowd rendering. Our approach renders high-quality, animated crowds in real-time (\cref{fig:crowd}). Our main contributions are as follows:

\begin{itemize}[itemsep=0pt, parsep=0pt, topsep=0pt, partopsep=0pt]
    \item 
    %To the best of our knowledge, we are the first to use the 3D Gaussian Splatting for crowd rendering. We introduce CrowdSplat, a system that enables real-time, high-quality rendering of animated crowds.
    A novel 3D Gaussian Splatting method for real-time rendering of realistic animated crowds (CrowdSplat).
    \item 
    %We enhance the Gaussian avatar method by sharing common Gaussian splat attributes for duplicate characters. The optimized CUDA program reduces memory usage which enables the rendering of large crowds.
    We optimize the CUDA implementation of Gaussian Avatar \cite{Hu2023GaussianAvatarTR} such that common 3D Gaussians attributes for duplicate characters are shared to reduce memory usage for large crowds.
    \item 
    %We improve the efficiency of Gaussian crowd rendering by implementing a LoD approach, using varying numbers of 3D gaussian splats at different distances to reduce the resources needed for real-time rendering without compromising rendering quality.
    A novel LoD approach, where the number of 3D Gaussians at different distances is varied to achieve real-time rendering without compromising rendering quality.
\end{itemize}

\begin{figure}[t]
    \centering
    \includegraphics[width=0.96\linewidth]{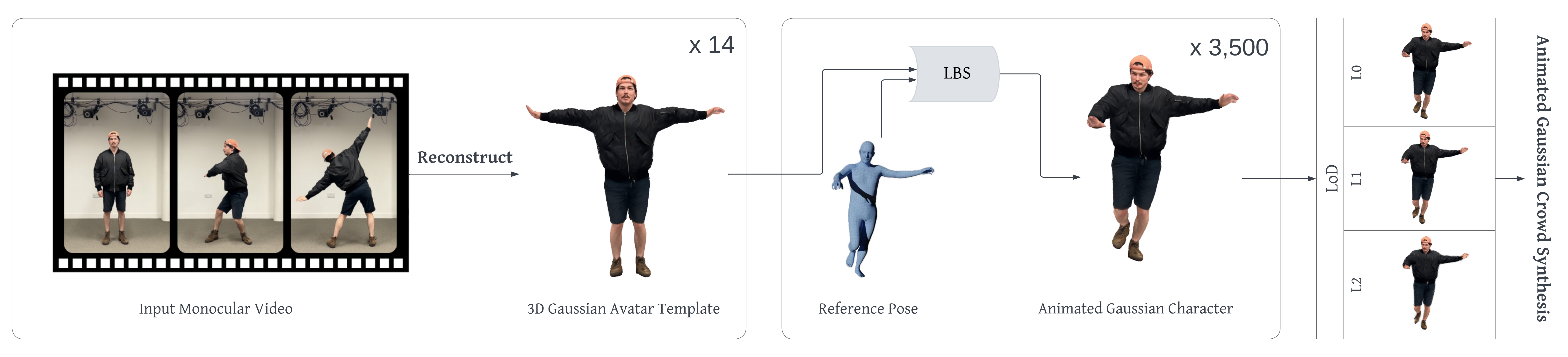}
%\includesvg[width=0.8\textwidth]{images/Blank.svg}
    \caption{\textbf{Overview of CrowdSplat.} The first stage combines the estimated SMPL body poses and images with a UV positional map. This process fits the 3D Gaussian attributes for each sampled point on the SMPL mesh template, reconstructing a Gaussian avatar template. The second stage uses Linear Blend Skinning (LBS) to animate multiple crowd characters, using an LoD technique for memory and rendering speed optimization. We reconstruct 14 avatar templates in the first stage and randomly duplicate these templates to 3,500 characters in the second stage.}
    \label{fig:pipeline}
\end{figure}
%\section{Background}
\vspace{-0.5cm}
\section{Background and Methods}
%\noindent
%\textbf{Crowd Rendering Methods:} 
\vspace{-0.35cm}
Previous approaches to real-time crowd rendering have used polygonal, point-based, image-based and hybrid representations \cite{CrowdSurvey2022}. Polygon-based techniques offer detailed geometric representations, but become expensive to compute with increasing realism. Point-based techniques are efficient but lack visual coherency for intricate details. Image-based techniques use texture-mapped impostors \cite{geo} to replace 3D characters, which reduces computation but is memory intensive and lacks realism in close views. 

%\paragraph{Crowd Rendering Methods:} Polygon-based techniques use meshes to represent characters, offering detailed geometric representations. However, creating high-quality simplified meshes for animation requires significant manual intervention and has high computational costs. Point-based techniques represent characters using point clouds, displayed with surface splatting primitives. This approach is efficient but struggles with visual coherence for intricate details. Image-based techniques use texture-mapped impostors to replace 3D characters, reducing computational load but lacking realism in close-up views and dynamic interactions.

%\vspace{-\baselineskip}
%\vspace{+0.1cm}
%\noindent
%\textbf{Scene Representation:} 
Neural rendering methods e.g., Neural Radiance Fields (NeRF) \cite{N1} can capture intricate details from sparse 2D images with an implicit representation, thus creating realistic images from limited data. 3D Gaussian Splatting (3DGS) \cite{Kerbl20233DGS}, an explicit representation, uses 3D Gaussians as primitives to represent scenes and manage complex effects with low computational cost. Building on this work, novel view synthesis methods have been proposed for static and dynamic scene reconstruction, such as clothed human movement \cite{Luiten2023Dynamic3G, Hu2023GaussianAvatarTR}, which is beneficial for real-time applications. 

%\paragraph{Neural Rendering and Scene Representation} Neural rendering uses neural networks to generate images, like Neural Radiance Fields (NeRF) \cite{N1} which create realistic images from limited data. Neural rendering captures intricate details from sparse 2D images with implicit representation. 3D Gaussian Splatting (3DGS) \cite{Kerbl20233DGS} is an explicit representation, which uses Gaussian distributions to represent scenes and manage complex effects with low computational cost. Researchers have proposed novel view synthesis for static scenes and dynamic scene reconstruction, such as clothed human movement \cite{Luiten2023Dynamic3G, Hu2023GaussianAvatarTR}, which is beneficial for applications requiring interactive frame rates, like gaming and virtual reality.

%\section{Method}
To render realistic results with detailed cloth wrinkles and animations, most Gaussian Avatar methods use a two-stage approach.  First, an avatar template is built on top of the SMPL template \cite{Loper2023SMPLAS}, which is a neural network model that depends on a specific subject's body shape and corresponding appearance. Next, a neural network is trained to predict the pose-dependent 3D Gaussian Splatting, which simulates subtle details such as cloth wrinkles. We also employ two stages, as shown in \cref{fig:pipeline}, where stage 1 is the same as in GaussianAvatar \cite{Hu2023GaussianAvatarTR}, whereas stage 2 differs in that a crowd is created from multiple reconstructed avatar templates with varying levels of detail, driven by multiple motion sequences.

We reconstruct 14 avatar templates for our crowd simulation, using monocular videos recorded from four subjects wearing different outfits and performing random movements for 2 minutes. GaussianAvatar \cite{Hu2023GaussianAvatarTR} uses a fixed opacity and rotation for each 3D Gaussian. We directly set these values in the CUDA program. Moreover, animated characters that share the same canonical template but differ in their pose can be further optimized. Since only their poses vary, only the 3D Gaussians' means require adjustment. Hence, we optimize the CUDA program by reusing all other Gaussian parameters across characters of the same template. This adjustment optimizes memory usage by approximately 11\% in a crowd of 5,000 characters.

To test our method, we build a crowd system containing 3,500 characters with 14 avatar templates placed randomly in a grid scene. The camera is static and three meters away from the crowd's first row, and 15 motion sequences are selected from the AMASS dataset \cite{amass} to animate the crowd. For level of detail, we use templates with 202k Gaussians for characters within 5 meters of the camera, 12k Gaussians from 5 to 10 meters, and 3k Gaussians beyond 10 meters. In this setup, we achieve a rendering speed of 31 frames per second for 3,500 characters, using the NVIDIA RTX 4090 GPU and Intel i9-13900 CPU.
% \pagebreak

\begin{figure}[tb]
    \centering
    % \includesvg[width=0.65\textwidth]{images/Qualitative.svg}
    \includegraphics[width=0.65\linewidth]{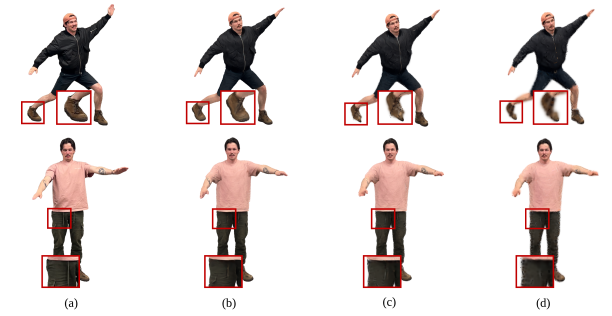}
    \caption{Qualitative rendered image comparison for different resolutions. (a) Ground truth test image, (b) Rendered image with 202,738 Gaussians, (c) Rendered image with 12,661 Gaussians, (d) Rendered image with 3,176 Gaussians.}
    \label{fig:QE}
\end{figure}

\section{Evaluation, Conclusions and Future Work}
\begin{figure}[t]
    \centering
    % \includesvg[width=0.65\textwidth]{images/plot.svg}
    \includegraphics[width=0.65\linewidth]{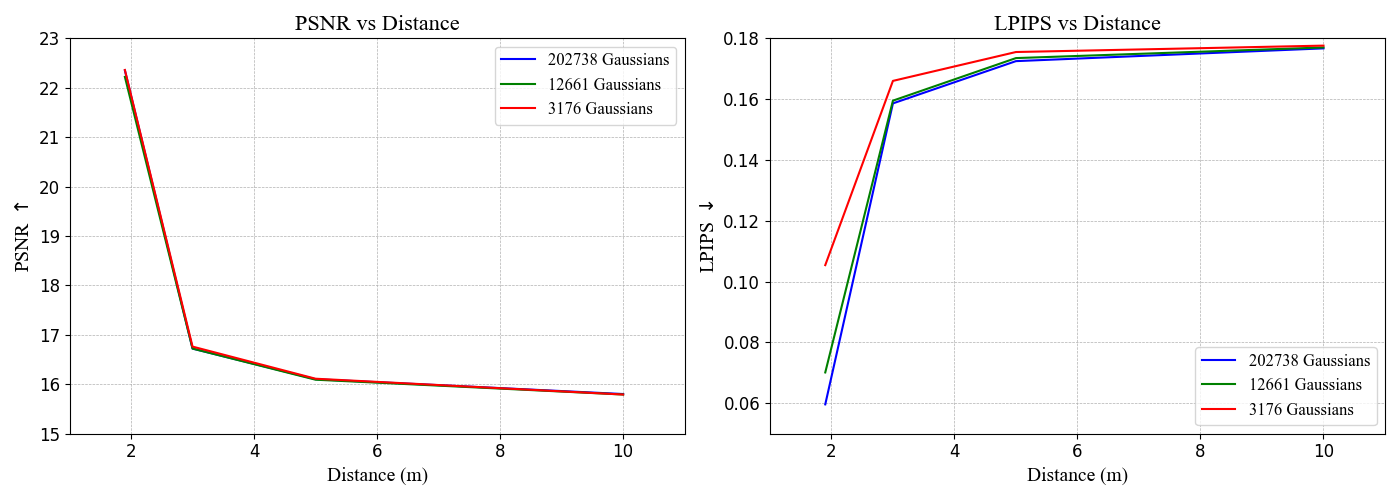}
    \caption{Quantitative results of rendering various numbers of Gaussians at different distances: The LPIPS metrics  \cite{Zhang2018TheUE} indicate that, within 5 meters, the quality of 202,738 and 12,661 Gaussians is higher than for 3,176 Gaussians, and similar at 10 meters, thus demonstrating the potential of CrowdSplat for Level of Detail (LoD) rendering.}
    \label{fig:result}
\end{figure}

%\paragraph{Quantitative Evaluation:} 
\cref{fig:QE} provides a qualitative comparison of images rendered with varying numbers of 3D Gaussians, while keeping the distance between the character and the camera at 1.9 meters. For quantitative comparisons of rendering quality, we employ two metrics: Peak Signal-to-Noise Ratio (PSNR) and Learned Perceptual Image Patch Similarity (LPIPS) \cite{Zhang2018TheUE}. Around 150 test poses are extracted from untrained frames of the original videos. We place 3k, 12k, and 202k avatar templates at distances of 1.9 meters, 3 meters, 5 meters, and 10 meters from the virtual camera. The resulting rendering quality metrics can be seen in \cref{fig:result}. The LPIPS metric, which better mirrors human perception, reveals that varying the number of 3D Gaussians makes a significant difference at 1.9 meters. However, this difference is not noticeable at 3 meters and 5 meters with 202k and 12k Gaussians. Beyond 10 meters, none of the lower-level avatar templates have a notable impact on the LPIPS metrics.

We calculate the GPU memory cost for different numbers of 3D Gaussians and quantities of duplicated characters, as shown in \cref{table:GPU Memory Comparison}. According to the table, our CUDA program optimization can reduce memory usage by around 11\% for 100 characters using the 202k Gaussians template. Similarly, it can reduce memory usage by around 12\% for 1,000 characters using the 12k Gaussians template, and by around 11\% for 5,000 characters using the 3k Gaussians template.

Similar to the GPU memory comparison, we evaluate rendering speed for various numbers of Gaussians and 
%quantities of 
duplicated characters, as shown in \cref{table:FPS Comparison}. We measure the rendering speed for both static and animated crowds with a resolution of 1,280 x 720. Notably, the animated crowd requires significant processing to apply LBS to the positions of templates.

\begin{table}[tb]
\centering
\begin{minipage}{0.45\textwidth}
\centering
\scriptsize
\caption{GPU memory cost (in MiB) for different numbers of Gaussians and characters. $^{\dagger}$ is the memory optimization by the CUDA program.}
\begin{tabular}{|>{\centering\arraybackslash}p{2.06cm}|>{\centering\arraybackslash}c|>{\centering\arraybackslash}c|>{\centering\arraybackslash}c|>{\centering\arraybackslash}c|>{\centering\arraybackslash}c|}
\hline
\diagbox[width=2.5cm]{\textbf{\# Gaussians}}{\textbf{\# characters}} & 1 & 100 & 400 & 1,000 & 5,000 \\
\hline
202,738 & 651 & 7,929 & - & - & - \\
\hline
202,738$^{\dagger}$ & 651 & 7,075 & - & - & - \\
\hline
12,661 & 567 & 1,005 & 2,335 & 5,127 & 21,340 \\
\hline
12,661$^{\dagger}$ & 567 & 953 & 2,121 & 4,593 & 19,625 \\
\hline
3,176 & 557 & 675 & 993 & 1,695 & 5,995 \\
\hline
3,176$^{\dagger}$ & 557 & 639 & 937 & 1,541 & 5,329 \\
\hline
\end{tabular}
\label{table:GPU Memory Comparison}
% \end{table}

\end{minipage}%
\hspace{0.04\textwidth}
\begin{minipage}{0.45\textwidth}
\centering
\scriptsize
\caption{Rendering speed (in FPS) for different numbers of Gaussians and characters}
\begin{tabular}{|>{\centering\arraybackslash}p{2.42cm}|>{\centering\arraybackslash}c|>{\centering\arraybackslash}c|>{\centering\arraybackslash}c|>{\centering\arraybackslash}c|>{\centering\arraybackslash}c|}
\hline
\diagbox[width=2.8cm]{\textbf{\# Gaussians}}{\textbf{\# characters}} & 1 & 100 & 400 & 1,000 & 5,000 \\
\hline
202,738 (w/o motion) & 728 & 50 & 22.2 & - & - \\
\hline
202,738 (w/ motion) & 617 & 18.2 & - & - & - \\
\hline
12,661 (w/o motion) & 1,612 & 586 & 251 & 125 & 10.3 \\
\hline
12,661 (w/ motion) & 1,370 & 312 & 76 & 34 & 6.1 \\
\hline
3,176 (w/o motion) & 1,703 & 1,240 & 673 & 275 & 38.6 \\
\hline
3,176 (w/ motion) & 1,347 & 804 & 298 & 121 & 23.2 \\
\hline
\end{tabular}
\label{table:FPS Comparison}
\end{minipage}
\end{table}
% \vspace{-\baselineskip}
%\paragraph{ Qualitative Evaluation:}

%\section{Conclusions and Future Work}
In conclusion, we introduced CrowdSplat, a novel approach for real-time, high-quality crowd rendering using 3D Gaussian Splatting and LoD techniques. Our method demonstrates the capability to render dynamic crowds from real-world humans, enabling arbitrary animation while maintaining interactive rendering speed.

Our future research will focus on several areas to further enhance the capabilities of CrowdSplat. We plan to explore the integration of hybrid systems that combine 3D Gaussian Splatting with image-based impostor techniques to optimize rendering efficiency and increase crowd appearance variety. A comprehensive set of user studies will provide valuable insights into the perception of 3DGS LoD techniques, and help determine the optimal number of 3D Gaussians. Additionally, we aim to develop methods for generating crowd scenes from textual descriptions, thus enabling a more diverse crowd creation workflow.
 
\section*{Acknowledgments}
This work was conducted with the financial support of the Science Foundation Ireland Centre for Research Training in Digitally-Enhanced Reality (d-real) at Trinity College Dublin under Grant No. 18/CRT/6224.

\bibliographystyle{apalike}

\bibliography{imvip}

\end{document}